\documentclass{article} 
\usepackage[numbers,sort&compress,square]{natbib}
\usepackage{iclr2027_conference,times}

\usepackage{amsmath,amsfonts,bm}

\def\eqref#1{equation~\ref{#1}}

\def\1{\bm{1}}

\DeclareMathAlphabet{\mathsfit}{\encodingdefault}{\sfdefault}{m}{sl}
\SetMathAlphabet{\mathsfit}{bold}{\encodingdefault}{\sfdefault}{bx}{n}

\definecolor{iccvblue}{rgb}{0.21,0.49,0.74}
\usepackage[pagebackref,breaklinks,colorlinks,allcolors=iccvblue]{hyperref}

\usepackage{xspace}        
\usepackage{graphicx}      
\usepackage{booktabs}      
\usepackage{wrapfig}       
\usepackage{array}         
\usepackage{multirow}      
\usepackage{tabularx}      
\usepackage{microtype}     
\usepackage{textcomp}      
\usepackage{gensymb}       
\usepackage{url}
\usepackage{bm}            
\usepackage{enumitem}      
\usepackage{pifont}        

\usepackage[table]{xcolor} 
\usepackage{colortbl}      

\usepackage{tikz}
\usepackage{pgfplots}
\pgfplotsset{compat=1.18}

\usepackage[switch]{lineno}

\usepackage{minitoc}       
\usepackage{tocloft}       
\usepackage[toc,page,header]{appendix} 

\usepackage[accsupp]{axessibility} 
 \usepackage{tcolorbox}          

\usepackage[capitalize]{cleveref}
\crefname{section}{Sec.}{Secs.}
\Crefname{section}{Section}{Sections}
\Crefname{table}{Table}{Tables}
\crefname{table}{Tab.}{Tabs.}

\usepackage{caption}

\definecolor{bestgreen}{RGB}{153,200,76}
\definecolor{worstred}{RGB}{192,0,0}

\definecolor{cbad}{HTML}{FFD0D0}
\definecolor{cmedium}{HTML}{FFF0D0}
\definecolor{cgood}{HTML}{90C060}

\newlength\savewidth

\definecolor{GreenColor}{rgb}{0.137,0.573,0.565}
\definecolor{DeltaColor}{rgb}{0.039,0.73,0.71}
\definecolor{SigmaColor}{rgb}{0.98,0.45,0.0}
\definecolor{AlphaColor}{rgb}{0,0,0.8}
\definecolor{BetaColor}{rgb}{0.8,0,0.8}
\definecolor{GammaColor}{rgb}{0.514,0.34,0.224}
\definecolor{EpsilonColor}{rgb}{0.353,0.725,0.906}
\definecolor{PurpleColor}{HTML}{9839ff}
\definecolor{RedColor}{rgb}{0.949,0.275, 0.224}
\definecolor{citecolor}{HTML}{0071bc}

\definecolor{deepred}{HTML}{940000}
\definecolor{cvprblue}{rgb}{0.21,0.49,0.74}
\hypersetup{linkcolor=deepred,urlcolor=cvprblue,citecolor=[rgb]{0.4,0.15,0.95}}

\newcommand{\longtitle}{Anatomically Plausible Human Image Generation via Synthetic Localized Preferences}\xspace           
\newcommand{\ourtitle}{\longtitle} 

\newcommand{\suppl}{\textcolor{magenta}{\emph{Sup.Mat.}}\xspace}

\definecolor{OrangeColor}{RGB}{255,140,0}          

\title{\ourtitle}

\author{Bao Li$^{1,2}$\thanks{This work was done during an internship at Westlake University.} \ Haichao Shi$^{1}$  \ Xiaoyu Zhang$^{1}$ \ Yuliang Xiu$^{2\dagger}$ \ Zhen Liu$^{3}$\thanks{Corresponding Author.} \\
$^{1}$Institute of Information Engineering, CAS \quad $^{2}$Westlake University \\ $^{3}$The Chinese University of Hong Kong, Shenzhen \\
{}
\\
Project Page: \url{https://britton-li.github.io/ASAP}}

\iclrfinalcopy 
\begin{document}
\doparttoc               
\faketableofcontents     


\maketitle

\lhead{Preprint.}

\vspace{-3mm}
\begin{abstract}
Large-scale text-to-image foundation models have substantially improved the visual realism of generated human images.
However, their outputs may still exhibit anatomical artifacts in structurally complex regions.
Preference-based alignment offers an architecture-preserving approach to improving anatomical plausibility.
In practice, conventional preference pairs typically compare independently generated images whose differences extend beyond anatomy, making it unclear whether anatomical quality drives the preference.
To provide direct anatomical supervision, we present Alignment via Synthetic Anatomical Preferences (ASAP), a data-centric framework built on synthetic localized preferences.
ASAP constructs synthetic localized preferences by using verified anatomically plausible images as preferred anchors and creating dispreferred images via localized degradation, while largely preserving prompt semantics and non-target visual content.
This pipeline yields the Human Anatomical Preference (HAP) dataset, comprising over 10K curated pairs with region masks and anatomical error metadata.
To exploit their spatial attribution while preserving overall image quality, we adapt Direct Preference Optimization (DPO) with a region-aware preference gap that emphasizes modified regions and a margin-bounded regression objective that drives the gap toward a finite target.
We further introduce HAF-Bench, a benchmark for systematic evaluation of anatomical fidelity across diverse scenarios. 
Extensive experiments across diffusion and flow-matching backbones demonstrate that ASAP consistently reduces anatomical errors while maintaining overall image quality.
\end{abstract}


\vspace{-2mm}
\section{Introduction}
\label{sec:intro}

Human image generation has made remarkable progress in visual quality, prompt alignment, and stylistic diversity with the development of text-to-image foundation models~\citep{podell2023sdxl,esser2024scaling,flux2024,wu2025qwen}. 
However, anatomical artifacts can still arise in challenging cases, ranging from malformed hands and incorrect finger counts to distorted facial structures~\citep{ju2023humansd, li2024cosmicman, wang2025fairhuman}. Such errors are particularly detrimental because human observers are highly sensitive to subtle deviations in faces, hands, and body structures. 
Existing approaches address this problem through human-centric training data, structural conditions, part-specialized modules, or inference-time refinement~\citep{li2024cosmicman,zhu2024mole,fang2024humanrefiner,wang2025realishuman}. While effective, these solutions often require task-specific components, additional control signals, or inference-time correction. Preference-based alignment offers a complementary, architecture-preserving alternative for teaching pretrained generators which anatomical structures should be preferred and which failures should be avoided.

The effectiveness of preference-based alignment depends critically on the supervision encoded in its preference pairs. 
Conventional text-to-image preference pairs typically compare independently generated images from the same prompt and subsequently ranked by users, annotators, reward models, or vision-language models~\citep{xu2023imagereward,kirstain2023pick,hpsv2,ma2025hpsv3}. 
Such pairs are well suited for learning holistic preferences such as aesthetics and prompt alignment, but are not constructed to target specific failure modes.
Human-focused approaches such as HG-DPO~\citep{na2025boost} narrow the data domain to human images, but their paired images may still differ across multiple semantic and visual attributes.
Consequently, the preference label is not determined by anatomical quality, limiting its effectiveness as anatomy-specific supervision.

\begin{figure}[tb]
  \centering
  \vspace{-12mm}
  \makebox[\textwidth][c]{\includegraphics[width=1.0\linewidth]{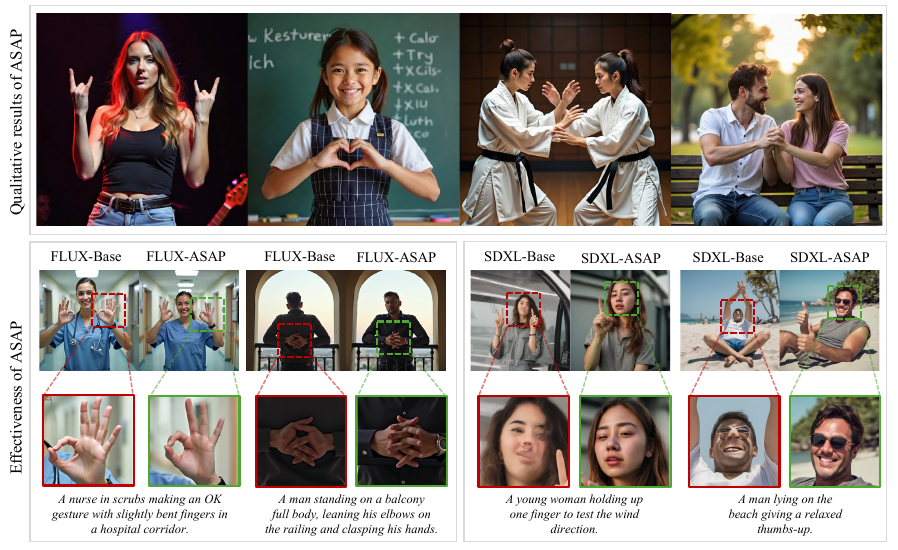}}
  \vspace{-6mm}
  \caption{\textbf{Effectiveness of ASAP.} Foundation models such as FLUX and SDXL still struggle with complex human anatomy, ASAP significantly improves anatomical plausibility.
  }
  \vspace{-6mm}
  \label{fig:first_vis}
\end{figure}

To address this problem, we present \emph{Alignment via Synthetic Anatomical Preferences (ASAP)}, an anatomy-focused preference-alignment framework built on synthetic localized preferences. 
Rather than pairing independently generated images, we curate human-verified synthetic images with plausible anatomy as preferred anchors and construct corresponding negatives by introducing localized anatomical errors in image or skeleton space. 
Since each negative is constructed from its preferred anchor, the pair largely preserves prompt semantics and non-target visual content, making the localized anatomical quality the principal source of preference. 
Using this pipeline, we construct the Human Anatomical Preference (HAP) dataset, comprising over 10K curated preference pairs with region masks and anatomical error metadata. 
Although synthetic localized preferences provide clear anatomy-specific supervision, standard DPO does not exploit their spatial attribution and continues to reward larger preference gaps. 
Our experiments suggest that excessive preference gaps may compromise overall image quality despite improvements in anatomical plausibility. 
We therefore adapt DPO with a mask-aware, margin-bounded objective that emphasizes the modified regions and drives the preference gap towards a finite target margin. 
For systematic evaluation, we further introduce HAF-Bench, a benchmark comprising 500 prompts balanced across five anatomical categories, together with a standardized evaluation protocol and complementary absolute and relative metrics. 

Our contributions are summarized as follows: (1) We introduce synthetic localized preferences for anatomical supervision and construct HAP, a dataset of over 10K pairs derived from human-verified synthetic anchors through localized degradations, with region masks and error metadata. (2) We propose a region-aware, margin-bounded preference objective that emphasizes modified regions while constraining excessive preference separation, improving the balance between anatomical plausibility and overall image quality. (3) We establish HAF-Bench, a 500-prompt benchmark for systematically evaluating anatomical plausibility. Experiments across diffusion- and flow-based generators demonstrate that ASAP improves anatomical plausibility over the pretrained generators and achieves a better balance between anatomical plausibility and global image quality.
\vspace{-2mm}



\section{Related Work}
\vspace{-1mm}
\subsection{Human Image Generation}
Human image generation is a central task in modern image synthesis, with broad applications in digital content creation and virtual humans~\citep{jia2024human}. Despite substantial advances in visual realism, modern text-to-image models~\citep{podell2023sdxl, flux2024, wu2025qwen} still produce artifacts in structurally complex regions, particularly hands, faces, and articulated body parts. Existing approaches address these failures by incorporating human-specific priors at different stages of the generation. At training time, HyperHuman and CosmicMan leverage large-scale human-centric data with detailed structural or semantic annotations, while HumanSD uses pose conditions to constrain body layout and articulation~\citep{liu2023hyperhuman,li2024cosmicman, ju2023humansd}. MoLE and FairHuman instead focus on anatomically sensitive regions through specialized experts and spatial supervision~\citep{zhu2024mole,wang2025fairhuman}. Other methods operate after generation: HumanRefiner combines pose-guided regeneration with localized inpainting, whereas RealisHuman regenerates malformed human parts and integrates them into the original image~\citep{fang2024humanrefiner,wang2025realishuman}. Our work explores a complementary, data-centric alignment approach that improves anatomical plausibility without modifying the generator architecture or inference pipeline.

\begin{figure}[t]
  \centering
  \vspace{-5mm}
  \includegraphics[width=0.95\linewidth]{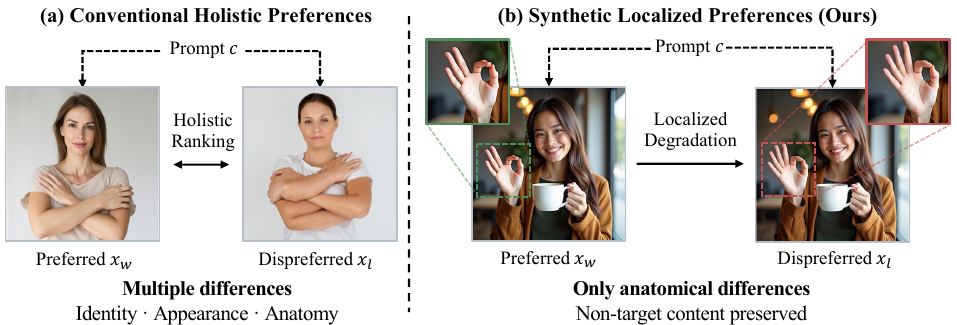}
  \vspace{-1mm}
  \label{preferences}
  \caption{\textbf{Conventional holistic preferences vs.\ synthetic localized preferences.} Conventional pairs may differ across attributes, obscuring the source of preference. ASAP constructs synthetic localized preferences through localized degradations, providing anatomy-specific preference signals.}
  \vspace{-5mm}
\end{figure}

\subsection{Alignment with Human Preference}
As foundation models are typically trained with huge and diverse unlabeled datasets~\citep{raffel2020exploring, penedo2023refinedweb}, the distributions they capture generally deviate from what humans prefer. In image generation, such misalignment is often observed as inferior aesthetics, text-image misalignment, image artifacts~\citep{schuhmann2022laion, kirstain2023pick, liang2024rich}. This misalignment is much more severe when it comes to human image generation since humans can easily identify visual artifacts of human bodies in generated images~\citep{mori2012uncanny}. 
Preference alignment is commonly performed through reinforcement learning with human feedback (RLHF), where a learned reward guides optimization, or direct preference optimization (DPO), which directly learns from preference pairs~\citep{ouyang2022training,rafailov2023direct}.
Originally proposed for large language models, DPO has been adapted to diffusion and flow-matching models~\citep{wallace2024diffusion, liu2026improving}.
However, DPO adjusts the likelihood of all pixels in an image uniformly, which can be inefficient since human preferences often depend on only a small subset of regions. Recent patch- and region-aware alignment methods use local feedback or spatial constraints~\citep{huang2025patchdpo,xing2025focus,wang2025harness}. 
For anatomical alignment, our objective emphasizes modified regions using degradation masks and drives the preference gap toward a finite target, improving anatomical plausibility while mitigating global image-quality degradation.

\subsection{Preference Data Construction for Visual Alignment}
Visual preference datasets commonly compare images independently generated for the same prompt and collect image-level judgments from users or annotators~\citep{wu2023human,kirstain2023pick,xu2023imagereward,ma2025hpsv3}. Such pairs are effective for learning holistic preferences, but may differ across multiple semantic and visual attributes, obscuring the source of each preference. HG-DPO narrows the data domain and improves preferred-sample quality using real human images, yet retains a holistic relationship between the paired samples~\citep{na2025boost}. 
Recent studies move toward reducing non-target differences by constructing closely matched preference pairs.
D-Fusion improves visual consistency for prompt alignment, while Di3PO, RealAlign, and LocalDPO develop task-specific preference construction strategies for text rendering, general visual alignment, and video generation, respectively~\citep{hu2025d,reddy2026di3po,chen2026preference,huang2026mind}. 
Within this line of work, ASAP address how to construct preference supervision for anatomical failures, which are local in appearance yet structured in pose and connectivity.


\section{Preliminaries}

\noindent \textbf{Diffusion and Flow-Matching Models.}
The prevalent image generation models are mostly built with diffusion models
(\textit{e.g.}, Stable Diffusion~\citep{rombach2022high}) and flow-matching
models (\textit{e.g.}, FLUX~\citep{flux2024}). 
Both can be described through time-dependent regression objectives.
Let $\mathbf{x}_1$ denote a clean image, $\mathbf{x}_0 \sim \mathcal{N}(0,I)$ an initial noise sample, and $\mathbf{x}_t$ their intermediate state at time $t$. Using $v_\theta$ to denote the model prediction, the model is trained with
the following loss:
\begin{align}
    \mathcal{L}(\theta) =
    \mathbb{E}_{t, \mathbf{x}_0, \mathbf{x}_1}
    \Bigl[
    w(t)\| v_\theta(\mathbf{x}_t, t) -
    u(\mathbf{x}_1, \mathbf{x}_0, t) \|^2
    \Bigr],
    \label{eq:flow}
\end{align}
where $u(\mathbf{x}_1, \mathbf{x}_0, t)$ is the target velocity field induced
by the reference path and $w(t)$ is a weighting factor depending on the model
design (often set to $1$). For instance, in diffusion models,
$
u(\mathbf{x}_t,\mathbf{x}_0,t)
=
\dot{\alpha}_t \mathbf{x}_0
+
\dot{s}_t
(\mathbf{x}_t - \alpha_t \mathbf{x}_0)/s_t,
$
where $\alpha(t)$ and $s(t)$ are the diffusion schedules; in a popular variant
called rectified flow~\citep{liu2022flow},
$
u(\mathbf{x}_1,\mathbf{x}_0,t) = \mathbf{x}_1 - \mathbf{x}_0.
$

\noindent \textbf{Direct Preference Optimization.} 
Direct Preference Optimization (DPO)~\citep{rafailov2023direct} aligns generative models with human preferences directly using preference datasets without any reinforcement learning process. 
Diffusion-DPO~\citep{wallace2024diffusion} adapts this principle to visual generative models by utilizing the model training loss as a proxy for log-likelihoods and building an approximate DPO loss.
Specifically, given a condition $\mathbf{c}$ and a preference pair $(\mathbf{x}_w, \mathbf{x}_l)$, the approximate preference gap is defined as~\citep{ziv2025mr, liu2026improving}:
\begin{equation}
\delta_{\theta}(\mathbf{x}_w, \mathbf{x}_l, \mathbf{c}) \approx - \Big( \mathcal{L}^{\theta}(\mathbf{x}_w) - \mathcal{L}^{\theta}(\mathbf{x}_l) - \big( \mathcal{L}^{\text{ref}}( \mathbf{x}_w) - \mathcal{L}^{\text{ref}}( \mathbf{x}_l) \big) \Big),
\end{equation}
where $\mathcal{L}$ implicitly conditions on $\mathbf{c}$ for brevity. Under the assumption of Bradley-Terry preference model~\citep{bradley1952rank, rafailov2023direct}, the so-called Diffusion-DPO loss is
\begin{equation}
\mathcal{L}_{\text{DPO}}(\theta) = -\mathbb{E}_{(\mathbf{x}_w, \mathbf{x}_l, \mathbf{c}) \sim \mathcal{D}} \big[ \log \sigma(\beta \delta_{\theta}(\mathbf{x}_w, \mathbf{x}_l, \mathbf{c})) \big],
\label{eq:dpo}
\end{equation}
where $\sigma$ is the sigmoid function and $\beta$ controls the deviation from the reference model. Flow-matching models can be similarly fine-tuned with it~\citep{liu2026improving} to yield satisfactory results.

\section{Methods}
\begin{figure}[t]
  \centering
  \vspace{-5mm}
  \includegraphics[width=1.0\linewidth]{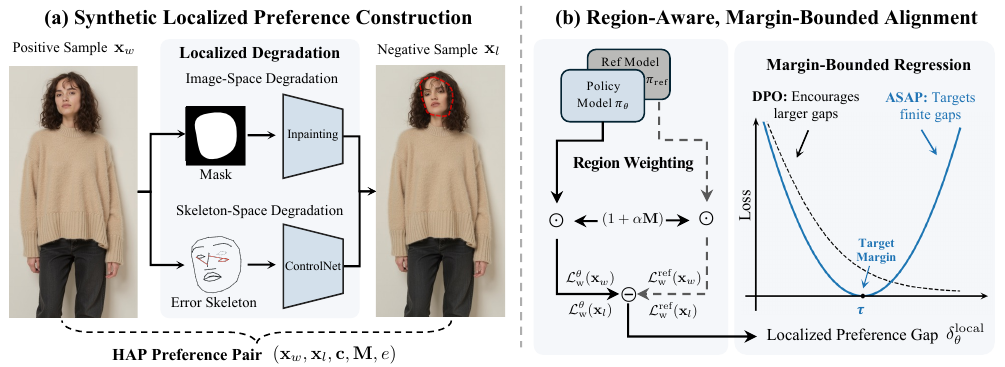}
  \vspace{-5mm}
  \caption{\textbf{Overview of the ASAP.} 
  \textbf{Phase I (left): Synthetic localized preference construction.} Starting from a verified preferred anchor $\mathbf{x}_w$, we introduce localized anatomical errors through image-space inpainting or skeleton-space perturbation to construct the corresponding negative $\mathbf{x}_l$.
  \textbf{Phase II (Right): Region-aware and margin-bounded preference optimization} . The region mask defines the spatial weighting $1+\alpha\mathbf{M}$ used to compute the localized preference gap. Rather than rewarding larger gaps as in standard DPO, our margin-bounded objective regresses the gap toward a finite target $\tau$, emphasizing the modified regions while mitigating over-optimization.}
  \vspace{-2mm}
  \label{pipeline}
\end{figure}

As illustrated in \cref{pipeline}, ASAP comprises two stages. We first construct the HAP dataset through synthetic localized preferences and then adapt DPO with mask-aware weighting and a finite target margin to exploit the spatial attribution provided by these pairs.

\subsection{Construction of Synthetic Localized Preferences}
\label{sec:synthesis}

\paragraph{Preferred Sample Construction.}
We first construct a pool of human-verified synthetic images with plausible anatomy as preferred samples. Instead of using existing datasets in the domain of human image generation, we adopt a pose-conditioned generation pipeline to create synthetic images for this dataset, so as to avoid the domain gap between photorealistic images and synthetic images during preference learning. Given a source human image, we extract its pose skeleton $\mathbf{p}$ using DWPose~\citep{yang2023effective} and use a vision-language model~\citep{comanici2025gemini} to refine the corresponding prompt into a detailed caption $\mathbf{c}_{\mathrm{aug}}$. These conditions guide a pretrained FLUX-ControlNet~\citep{flux-cn-union-pro-2} to generate candidate preferred samples $\mathbf{x}_w \sim G(\mathbf{c}_{\mathrm{aug}},\mathbf{p})$, from which annotators remove images with implausible anatomy. Each retained sample subsequently serves as the anchor for constructing its dispreferred counterpart.

\paragraph{Dispreferred Sample Construction.}
For each preferred sample $\mathbf{x}_w$, we construct a corresponding dispreferred sample $\mathbf{x}_l$ by introducing a localized anatomical error in either image or skeleton space. For image-space gradation, GroundingSAM2~\citep{ren2024grounded} extracts a binary mask $\mathbf{M}$ for the target region, which is regenerated using an inpainting model conditioned on an anatomical-error instruction. For skeleton-space degradation, we perturb $\mathbf{p}$ by modifying anatomical keypoints or connections to obtain $\mathbf{p}_{\mathrm{error}}$, and regenerate $\mathbf{x}_l$ using the same caption and FLUX-ControlNet. The affected region and error type are retained as the region mask and anatomical error metadata.

\begin{wraptable}{r}{0.4\textwidth}
\vspace{-0mm}
\centering
\caption{\textbf{Spatial locality of HAP pairs.}}
\label{locality}
\small
\setlength{\tabcolsep}{8pt}
\renewcommand{\arraystretch}{1.05}
\begin{tabular}{@{}lcc@{}}
\toprule
Region & SSIM & LPIPS \\
\midrule
Masked & 0.5315 & 0.3446 \\
Non-masked & 0.9827 & 0.0109 \\
\bottomrule
\end{tabular}
\vspace{-3mm}
\end{wraptable}

\paragraph{HAP Dataset Curation.} 
Applying the two construction strategies produces approximately 25K candidate preference pairs. We employ a panel of 11 human annotators to manually screen these pairs according to three criteria: (1) the preferred sample exhibits plausible anatomy and satisfactory overall quality, (2) the dispreferred sample contains a clearly identifiable anatomical error, and (3) prompt semantics and non-target visual content remain sufficiently consistent within the pair. A pair is retained only when all three criteria are satisfied. After screening, the Human Anatomical Preference (HAP) dataset contains over 10K curated tuples $(\mathbf{x}_w,\mathbf{x}_l,\mathbf{c},\mathbf{M},\mathbf{e})$, where $\mathbf{e}$ denotes the anatomical error metadata. Further curation details are provided in~\suppl's~\cref{app:hap}. \cref{locality} confirms this locality, with substantially higher SSIM and lower LPIPS outside the degradation masks.

\subsection{Region-Aware and Margin-Bounded Preference Optimization}
\label{sec:apo}
Each synthetic localized preference pair includes a region mask $\mathbf{M}$ that identifies where the anatomical error is introduced. Standard image-level DPO neither explicitly exploits this spatial attribution nor specifies a finite preference gap at which the desired correction is considered sufficient. We therefore adapt DPO with a region-aware preference gap and a margin-bounded objective.

\paragraph{Localized Preference Gap.} 
Given an image $\mathbf{x}$, its text condition $\mathbf{c}$, and the corresponding region mask $\mathbf{M}$, we define the mask-weighted training loss as 
\begin{equation}
\mathcal{L}_{\text{w}}(\theta; \mathbf{x}, \mathbf{c}, \mathbf{M})
= \mathbb{E}_{t,\mathbf{x}_0} \Bigl[
\big\| \sqrt{1+\alpha\mathbf{M}} \odot
[v_\theta(\mathbf{x}_t,t,\mathbf{c})-(\mathbf{x}_1-\mathbf{x}_0)]
\big\|_2^2
\Bigr],
\end{equation}

where $\odot$ denotes element-wise multiplication, and $\alpha$ controls the weighting strength. The factor 1+$\alpha \mathbf{M}$ retains unit weight outside the mask and assigns additional weight to the target region.

\noindent The localized preference gap is thus ($\mathcal{L}_{w}^{\theta}(\mathbf{x})$ stands for $\mathcal{L}_{w}(\theta; \mathbf{x}, \mathbf{c})$):
\begin{equation}
    \delta_{\theta}^{\text{local}}(\mathbf{x}_w, \mathbf{x}_l, \mathbf{c}, \mathbf{M}) \approx - \Big( \mathcal{L}_{\text{w}}^{\theta}(\mathbf{x}_w) - \mathcal{L}_{\text{w}}^{\theta}(\mathbf{x}_l) - \big( \mathcal{L}_{\text{w}}^{\text{ref}}(\mathbf{x}_w) - \mathcal{L}_{\text{w}}^{\text{ref}}(\mathbf{x}_l) \big) \Big).
\end{equation}

\paragraph{Margin-Bounded Regression.} 
To constrain the localized preference gap to a finite target, we adopt the margin-constrained formulation to  the DPO variant for flow matching models ~\citep{azar2024general, liu2026improving} and define the following regression objective:
\begin{equation}
    \label{eq:apo_loss}
    \mathcal{L}_{\text{Bounded}}(\theta) = \mathbb{E}_{(\mathbf{x}_w, \mathbf{x}_l, \mathbf{c}, \mathbf{M}) \sim \mathcal{D}} \left[ \left( \delta^{\text{local}}_{\theta}(\mathbf{x}_w, \mathbf{x}_l, \mathbf{c}, \mathbf{M}) - \tau \right)^2 \right],
\end{equation}

where $\tau$ denotes the finite target margin. 
By penalizing deviations from $\tau$, the objective prevents excessive preference gaps.

\section{HAF-Bench}
\label{sec:bench}

We introduce HAF-Bench (Human Anatomical Fidelity Benchmark), a 500-prompt benchmark for evaluating anatomical plausibility in generated human images. Complementing prior human-image benchmarks designed for artifact localization or broad quality assessment, HAF-Bench targets structural integrity, articulation, and physically plausible interactions across diverse scenarios.

\paragraph{Prompts and Taxonomy.} 
The prompts are generated with an LLM and subsequently filtered to remove ambiguous descriptions.
HAF-Bench contains 100 prompts in each of five complementary categories: (1) Basic Anatomy, covering the integrity of individual body parts; (2) Semantic Gestures, covering meaning-specific hand gestures; (3) Self-Interaction, covering contact between body parts; (4) Object Interaction, covering human–object contact; and (5) Full Body, covering global pose and anatomical connectivity.

\paragraph{Evaluation Protocol.} 
We use Gemini-3-Pro as an automated judge under a fixed anatomy-focused instruction. For absolute evaluation, the judge assigns each generated image a Pass label if no anatomical artifact is detected and a Fail label otherwise. For relative evaluation, it compares two images generated from the same prompt and returns Win, Tie, or Lose according to their anatomical plausibility. 
All methods use identical instructions; judge prompts and protocol details are provided in~\suppl's~\cref{app:evaluation}.
On the full evaluation, Gemini-3-Pro agrees with human judgments in 82\% of cases, supporting its use as a scalable anatomy evaluator.

\paragraph{Metrics.}
For absolute evaluation, we report the Anatomical Error Rate (AER), defined as the proportion of generated images judged to contain anatomical artifact. Let $\mathcal{D}_{\mathrm{HAF}}$ denote the set of evaluated images generated from HAF-Bench prompts, and let $\mathcal{S}_{\mathrm{VLM}}(\mathbf{x}) \in \{\text{Pass},\text{Fail}\}$ denote the judgment for image $\mathbf{x}$. AER is computed as
\begin{equation} \text{AER} = \frac{1}{|\mathcal{D}_{\text{HAF}}|} \sum_{\mathbf{x} \in \mathcal{D}_{\text{HAF}}} \mathbb{I}(\mathcal{S}_{\text{VLM}}(\mathbf{x}) = \text{Fail}), \end{equation}
where each image contributes a single binary outcome regardless of the number or severity of its artifacts. A lower AER indicates better anatomical plausibility.

For relative evaluation, we define the Relative Superiority Index (RSI) to summarize pairwise outcomes while retaining ties:
\begin{equation}
\mathrm{RSI}
=
\frac{N_{\mathrm{win}}+N_{\mathrm{tie}}}
{N_{\mathrm{lose}}+N_{\mathrm{tie}}},
\label{eq}
\end{equation}
where $N_{\mathrm{win}}$, $N_{\mathrm{tie}}$, and $N_{\mathrm{lose}}$ denote the corresponding outcomes from the evaluated model's perspective. RSI is greater than $1$ exactly when wins outnumber losses. Because ties enter the numerator and denominator symmetrically, tie-heavy comparisons produce values closer to $1$.


\section{Experiments}
\label{sec:experiments}

\subsection{Experimental Settings}
\paragraph{Base Models.}
We evaluate ASAP on two text-to-image backbones: FLUX.1-dev \citep{flux2024}, based on flow matching, and SDXL \citep{podell2023sdxl}, based on diffusion. Together, they evaluate the applicability of ASAP across two widely used generative formulations and model scales.

\paragraph{Baselines.}
We compare supervised fine-tuning and preference optimization methods. The supervised baselines include:
(1) \textit{SFT}, which fine-tunes each backbone using only preferred HAP samples; and
(2) \textit{FairHuman}$^\dagger$~\citep{wang2025fairhuman}, which jointly optimizes global, hand, and face objectives.
The preference optimization baselines include:
(3) \textit{HG-DPO}$^\dagger$~\citep{na2025boost}, which constructs preference pairs by selecting higher- and lower-reward images from generated candidates for each prompt; and
(4) \textit{ASAP-DPO}, which applies vanilla DPO to the same HAP pairs without the localized preference gap or margin-bounded regression.
\textit{ASAP} denotes our complete framework. Further baseline and implementation details are provided in~\suppl's~\cref{app:baselines}.

\paragraph{Evaluation Benchmarks and Metrics.}
We evaluate anatomical plausibility on HAF-Bench using AER for absolute evaluation and RSI for pairwise comparison. RSI compares each adapted model against its corresponding base model, whose value is $1.00$ by definition.
To assess whether anatomical improvements preserve general visual quality and prompt alignment, we additionally report HPSv3~\citep{ma2025hpsv3}, HPSv2.1~\citep{wu2023human}, PickScore~\citep{kirstain2023pick}, ImageReward~\citep{xu2023imagereward}, and Aesthetic Score~\citep{schuhmann2022laion}. All methods use identical prompts and sampling settings, with each metric averaged over five seeds.

\paragraph{Implementation Details.}
All adapted models are fine-tuned using rank-$32$ LoRA~\citep{hu2022lora} for 3,000 steps with an effective batch size of $16$. For FLUX.1-dev, we use a learning rate of $5\times10^{-6}$. For ASAP, we set the spatial weighting coefficient to $\alpha=10$ and the target margin to $\tau=0.01$. Detailed hyperparameter settings for the SDXL are provided in~\suppl's~\cref{app:settings}.

\begin{table}[t]
    \centering
    \vspace{-6mm}
    \caption{\textbf{Quantitative comparison on HAF-Bench.}
    AER and RSI assess anatomy quality; the remaining metrics assess global
    image quality. Base provides the reference; adapted methods are grouped
    into supervised fine-tuning and preference optimization for each backbone.}
    \label{tab:main_results}
    \small
    \setlength{\tabcolsep}{4pt}
    \setlength{\arrayrulewidth}{0.35pt}
    \setlength{\aboverulesep}{2pt}
    \setlength{\belowrulesep}{2pt}
    \renewcommand{\arraystretch}{1.06}
    \definecolor{ASAPRow}{RGB}{253,243,254}
    \resizebox{\linewidth}{!}{%
    \begin{tabular}{l|cc|ccccc}
        \toprule[0.8pt]
        \multirow{2}{*}{\textbf{Method}}
        & \multicolumn{2}{c|}{\rule{0pt}{2.5ex}\textit{Anatomy Quality}}
        & \multicolumn{5}{c}{\textit{Global Image Quality}} \\
        \cmidrule(lr){2-3}\cmidrule(lr){4-8}
        & AER $\downarrow$
        & RSI $\uparrow$
        & HPS v3 $\uparrow$
        & HPS v2.1 $\uparrow$
        & PickScore $\uparrow$
        & ImageReward $\uparrow$
        & AesScore $\uparrow$ \\
        \midrule[0.35pt]

        \rule{0pt}{2.5ex}\textit{FLUX.1-dev:}
        & & & & & & & \\
        Base
        & 0.417 & 1.00 (Ref.) & 10.307 & \underline{0.303} & 22.797 & 1.037 & 5.793 \\

        \rule{0pt}{2.4ex}{\footnotesize\itshape\color{black!65}Supervised fine-tuning}
        & & & & & & & \\
        \hspace{0.6em}SFT
        & 0.391 & 1.03 & 10.320 & \underline{0.303} & 22.808 & 1.097 & \textbf{5.822} \\
        \hspace{0.6em}FairHuman$^{\dagger}$
        & 0.346 & 1.09 & \underline{10.324} & \textbf{0.304} & 22.792 & 1.093 & \underline{5.806} \\

        \rule{0pt}{2.4ex}{\footnotesize\itshape\color{black!65}Preference optimization}
        & & & & & & & \\
        \hspace{0.6em}HG-DPO$^{\dagger}$
        & 0.371 & 1.05 & 10.323 & 0.302 & \textbf{22.967} & \underline{1.098} & 5.772 \\
        \hspace{0.6em}\textbf{ASAP-DPO (Ours)}
        & \underline{0.293} & \underline{1.13} & 9.680 & 0.299 & 22.476 & 1.062 & 5.752 \\
        \rowcolor{ASAPRow}
        \hspace{0.6em}\textbf{ASAP (Ours)}
        & \textbf{0.289} & \textbf{1.17} & \textbf{10.360}
        & \textbf{0.304} & \underline{22.814} & \textbf{1.102} & 5.779 \\
        \midrule[0.35pt]

        \rule{0pt}{2.5ex}\textit{SDXL:}
        & & & & & & & \\
        Base
        & 0.873 & 1.00 (Ref.) & 6.371 & 0.277 & 22.205 & 0.737 & 5.709 \\

        \rule{0pt}{2.4ex}{\footnotesize\itshape\color{black!65}Supervised fine-tuning}
        & & & & & & & \\
        \hspace{0.6em}SFT
        & 0.820 & 1.23 & 7.233 & \underline{0.289} & \underline{22.369} & 0.770 & \textbf{5.766} \\
        \hspace{0.6em}FairHuman$^{\dagger}$
        & 0.791 & 1.42 & 7.319 & 0.286 & 22.341 & 0.759 & \underline{5.763} \\

        \rule{0pt}{2.4ex}{\footnotesize\itshape\color{black!65}Preference optimization}
        & & & & & & & \\
        \hspace{0.6em}HG-DPO$^{\dagger}$
        & 0.804 & 1.31 & \underline{7.368} & 0.287 & \textbf{22.512} & \textbf{0.781} & 5.750 \\
        \hspace{0.6em}\textbf{ASAP-DPO (Ours)}
        & \underline{0.775} & \underline{1.49} & 6.526 & 0.279 & 22.254 & 0.751 & 5.723 \\
        \rowcolor{ASAPRow}
        \hspace{0.6em}\textbf{ASAP (Ours)}
        & \textbf{0.764} & \textbf{1.58} & \textbf{7.450}
        & \textbf{0.291} & 22.364 & \underline{0.778} & 5.754 \\
        \bottomrule[0.8pt]
    \end{tabular}%
    }
    \vspace{-6mm}
\end{table}

\subsection{General Results}
\paragraph{Quantitative Analysis.}
\cref{tab:main_results} compares ASAP with supervised fine-tuning and preference-optimization baselines on HAF-Bench. Across both backbones, ASAP achieves the lowest AER and highest RSI while maintaining competitive global image-quality scores. 
Applying vanilla DPO to the same synthetic localized preferences (ASAP-DPO) already substantially reduces anatomical errors, demonstrating the effectiveness of HAP as preference supervision. However, this variant consistently degrades global quality.
ASAP retains these anatomical improvements while largely recovering the global metrics, indicating that the region-aware, margin-bounded objective better balances localized anatomical correction and global quality preservation.

\begin{figure}[!t]
  \centering
  \vspace{-7mm}
  \makebox[\textwidth][c]{\includegraphics[width=0.95\linewidth]{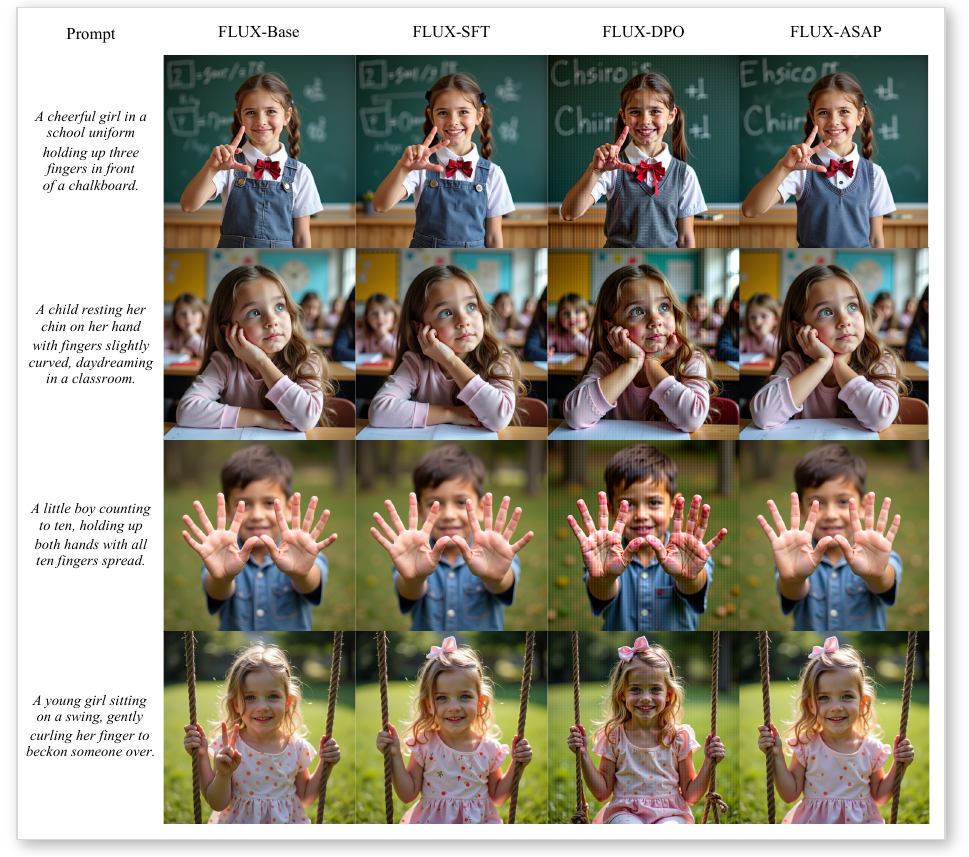}}
  \caption{\textbf{Qualitative comparison of fine-tuning strategies on FLUX.} 
  SFT uses the preferred images, ASAP-DPO applies vanilla DPO to the synthetic preference pairs, and ASAP further introduces region-aware, margin-bounded objective. ASAP corrects localized anatomical errors while better preserving non-target content and overall image quality.
  }
  \vspace{-5mm}
  \label{fig:ablation}
\end{figure}

\paragraph{Qualitative Analysis.} 
As shown in \cref{fig:first_vis}, ASAP consistently reduces anatomical artifacts in images generated by both FLUX and SDXL, particularly malformed hands and faces, while preserving the overall image quality.
To compare different fine-tuning strategies, \cref{fig:ablation} further presents samples generated by the base FLUX model, SFT, ASAP-DPO, and ASAP. The base model exhibits visible anatomical errors, which SFT often leaves unresolved. ASAP-DPO improves anatomical plausibility but at the cost of lower overall visual quality. In contrast, ASAP retains these anatomical improvements while better preserving global visual quality.

\subsection{Analysis and Ablation Studies}
\paragraph{Comparison of Preference Construction Strategies.}
To assess preference-pair construction, we compare four strategies under vanilla DPO objective: (1) \textit{Standard Pairwise Ranking}, which ranks two independently generated images for each prompt; (2) \textit{HG-DPO-Easy}~\citep{na2025boost}, which selects the highest- and lowest-scoring samples from a larger candidate pool; (3) \textit{Real-Anchor Localized Preferences}, which construct localized negatives from real human images; and (4) our \textit{Synthetic Localized Preferences}, constructed from human-verified synthetic images. 
All variants use the same prompt set, number of training pairs, and optimization settings, construction details are provided in~\suppl's~\cref{strategies}. 
As shown in \cref{tab:preference_strategy}, synthetic localized preferences achieve the lowest AER and highest RSI across both backbones. 
Their advantage over ranking-based strategies demonstrates the benefit of reducing non-target differences, while the comparison with real anchors suggests that matching the preference data to the model's generative domain improves anatomical alignment.

\paragraph{Ablation of the Alignment Objective.}
\cref{tab:objective_ablation} evaluates the two objective components while keeping the HAP dataset and training settings fixed. Using the localized preference gap alone achieves the lowest AER, showing that spatially focused supervision strengthens anatomical correction. Bounded regression instead substantially improves HPSv3 but produces more conservative anatomical correction. Combining both components improves AER over vanilla DPO while achieving the highest HPSv3, balancing between anatomical plausibility and overall generation quality.

\begin{table}[t]
\centering
\vspace{-9mm}

\begin{minipage}[t]{0.49\linewidth}
\centering
\caption{\textbf{Comparison of preference construction strategies.}
All variants use vanilla DPO with matched data and training settings.}
\label{tab:preference_strategy}

\scriptsize
\setlength{\tabcolsep}{3.0pt}
\begin{tabular}{@{}lcccc@{}}
\toprule
& \multicolumn{2}{c}{FLUX.1-dev}
& \multicolumn{2}{c}{SDXL} \\
\cmidrule(lr){2-3}
\cmidrule(lr){4-5}
Preference Strategy
& AER $\downarrow$ & RSI $\uparrow$
& AER $\downarrow$ & RSI $\uparrow$ \\
\midrule
Base
& 0.417 & 1.00
& 0.873 & 1.00 \\
\midrule
Standard Pairwise Ranking
& 0.415 & 0.98
& 0.823 & 1.13 \\

HG-DPO-Easy
& 0.371 & 1.05
& 0.804 & 1.31 \\

Real-Anchor Localized
& 0.356 & 1.08
& 0.783 & 1.43 \\

Synthetic Localized (Ours)
& \textbf{0.293} & \textbf{1.13}
& \textbf{0.775} & \textbf{1.49} \\
\bottomrule
\end{tabular}
\end{minipage}
\hfill
\begin{minipage}[t]{0.49\linewidth}
\centering
\caption{\textbf{Ablation of the alignment objective.}
All variants use the same HAP pairs with matched training settings on FLUX.1-dev.}
\label{tab:objective_ablation}

\scriptsize
\setlength{\tabcolsep}{5.0pt}
\renewcommand{\arraystretch}{1.08}
\begin{tabular}{@{}cccc@{}}
\toprule
\shortstack{Localized }
& \shortstack{Bounded}
& AER $\downarrow$
& HPSv3 $\uparrow$ \\
\midrule
$\times$ & $\times$
& 0.293 & 9.680 \\

$\checkmark$ & $\times$
& \textit{0.283} & \textit{9.717} \\

$\times$ & $\checkmark$
& \textit{0.337} & \textit{10.355} \\

$\checkmark$ & $\checkmark$
& 0.289 & 10.360 \\
\bottomrule
\end{tabular}

\vspace{1mm}
\parbox{0.94\linewidth}{
\footnotesize\raggedright
\textit{Note: The first row corresponds to vanilla DPO, and the last row to the complete ASAP objective.} 
}
\end{minipage}

\vspace{-2mm}
\end{table}

\begin{wrapfigure}{r}{0.5\textwidth}
  \vspace{-13mm} 
  \centering
  \includegraphics[width=0.95\linewidth]{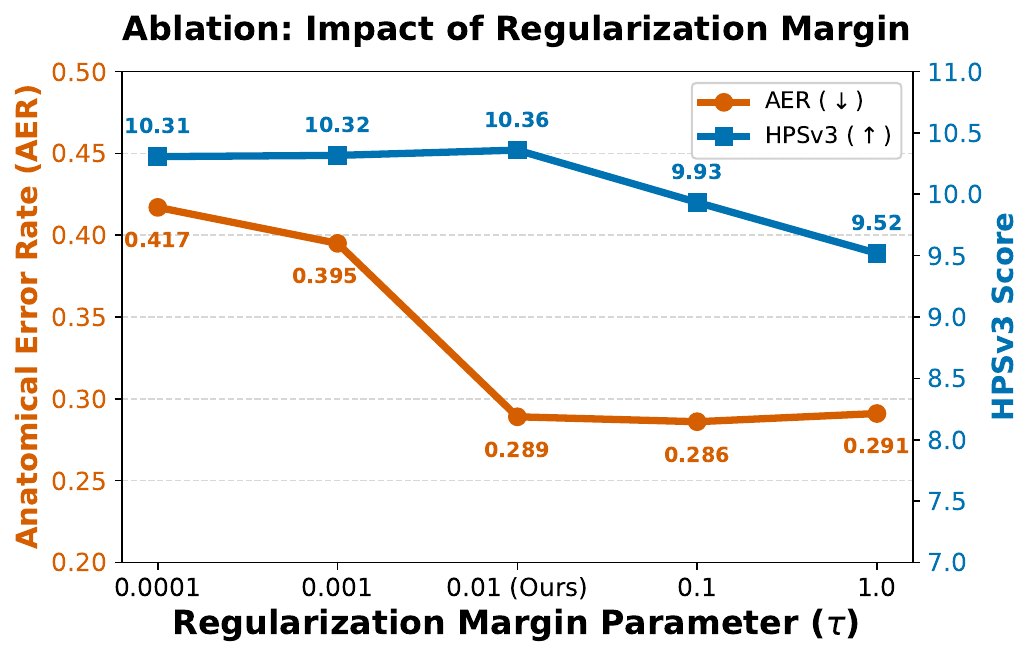}
  \vspace{-4mm}
  \caption{\textbf{Sensitivity to the Target Margin.}AER ($\downarrow$) and HPSv3 ($\uparrow$) across values of $\tau$. Small margins result in insufficient anatomical correction, whereas large margins reduce HPSv3 with little further improvement in AER. $\tau=0.01$ achieves the highest HPSv3 while maintaining near-minimal AER, providing the best trade-off.} 
  \label{fig:margin}
  \vspace{-12mm} 
\end{wrapfigure}

\paragraph{Sensitivity to the Target Margin.}
We examine how the target margin $\tau$ affects anatomical plausibility and overall generation quality, measured by AER and HPSv3, respectively. As shown in Fig.~\ref{fig:margin}, small margins ($\tau \leq 0.001$) constrain the preference gap too tightly, resulting in insufficient anatomical correction and higher AER. In contrast, large margins ($\tau \geq 0.1$) provide little further improvement in AER while substantially reducing HPSv3.
Among the evaluated settings, $\tau=0.01$ achieves the highest HPSv3 score of $10.36$ with a near-minimal AER of $0.289$, providing the best empirical balance between anatomical plausibility and overall generation quality.

\subsection{Human Evaluation}
\label{sec:human_evaluation}
\begin{figure}[tb]
  \centering
  \vspace{0mm}
  \makebox[\textwidth][c]
  {\includegraphics[width=0.8\linewidth]{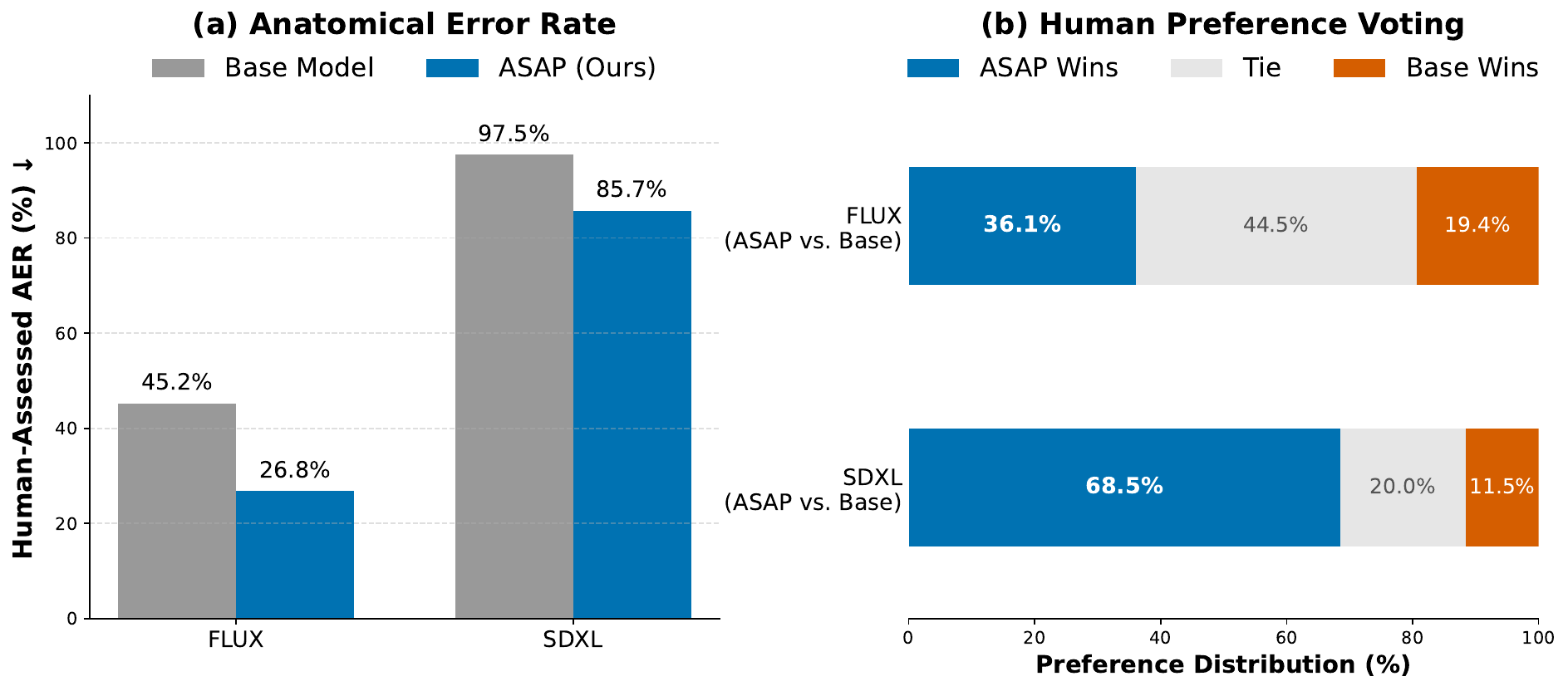}}
  \vspace{-6mm}
  \caption{  \textbf{Human Evaluation.} (a) Human-assessed AER for the base and ASAP-tuned models. ASAP reduces anatomical errors on both FLUX and SDXL. (b) Pairwise preference voting between ASAP and the base model. ASAP wins more frequently than the base model for both backbones.
  }
  \label{fig:human_evaluation}
  \vspace{-5mm}
\end{figure}

To assess whether the automated results align with human perception, we conduct human evaluation on the full HAF-Bench using the same absolute and pairwise protocols. As shown in Fig.~\ref{fig:human_evaluation}(a), ASAP reduces human-assessed AER for both FLUX and SDXL, confirming that its anatomical improvements are perceptible to human observers. Pairwise voting in Fig.~\ref{fig:human_evaluation}(b) further favors ASAP over the corresponding base models: ASAP wins more frequently than the base model for both backbones. Together, these results corroborate the automated evaluation and provide consistent human evidence for the effectiveness of ASAP.

\section{Conclusion}
\label{sec:conclusion}

In this paper, we present ASAP, a data-centric preference-alignment framework for improving anatomical plausibility in human image generation.
ASAP constructs synthetic localized preferences by introducing localized anatomical degradations into human-verified synthetic anchors, and couples them with a region-aware, margin-bounded objective that balances anatomical correction with overall visual quality.
We further introduce HAF-Bench for systematic evaluation across diverse anatomy-sensitive scenarios.
Experiments on flow-matching and diffusion backbones show that ASAP reduces anatomical errors while preserving overall image quality, with human evaluations corroborating these improvements.
These results demonstrate the value of anatomy-specific synthetic localized preferences for fine-grained alignment of text-to-image models.


\section*{Acknowledgements}
This work was supported by the National Natural Science Foundation of China
(No.~62602521), the Research Center for Industries of the Future (RCIF) at Westlake University, and the Westlake Education Foundation.


\bibliography{references}
\bibliographystyle{iclr2027_conference}

\clearpage
\newpage
\appendix

\addcontentsline{toc}{section}{Appendix} 
\renewcommand \thepart{} 
\renewcommand \partname{}
\part{\Large{\centerline{Appendix}}}
\parttoc

\newpage

\section{Use of Large Language Models}
We used a large language model to assist with copy editing---grammar checking,
wording suggestions, and minor style and clarity improvements---after the
scientific content, methodology, analyses, and conclusions had been written
by the authors.

\section{HAP Dataset Construction and Validation}
\label{app:hap}

\subsection{Data Curation Protocol}
\label{app:curation}
The localized degradation pipelines initially produce approximately 25K candidate preference pairs. A panel of 11 trained annotators independently screens the candidate pool, retaining a pair only when all three of the following criteria are satisfied:
\begin{itemize}[leftmargin=*, topsep=2pt, itemsep=2pt]
\item \textbf{Plausible Preferred Sample.} The preferred image $\mathbf{x}_w$ must exhibit plausible anatomy and satisfactory overall visual quality. Pairs are discarded if the original synthetic anchor contains perceptible anatomical artifacts.

\item \textbf{Identifiable Anatomical Degradation.} The dispreferred image $\mathbf{x}_l$ must contain a clearly identifiable anatomical error. Pairs with imperceptible modifications or without meaningful structural degradation are discarded.

\item \textbf{Preserved Non-Target Content.} Prompt semantics and visual content outside the degraded region must remain sufficiently consistent within the pair. Pairs exhibiting noticeable unintended changes in identity, composition, background, or other non-target regions are discarded.

\end{itemize}

After filtering, 10,652 pairs, corresponding to approximately 42.6\% of the candidate pool, form the final HAP dataset. Each retained tuple $(\mathbf{x}_w,\mathbf{x}_l,\mathbf{c},\mathbf{M},e)$ contains a preferred image, a dispreferred image, the corresponding caption, a degradation mask, and anatomical error metadata.

\subsection{Representative Failure Modes}
\begin{figure}[t]
  \centering
  \vspace{-2mm}
  \includegraphics[width=0.8\linewidth]{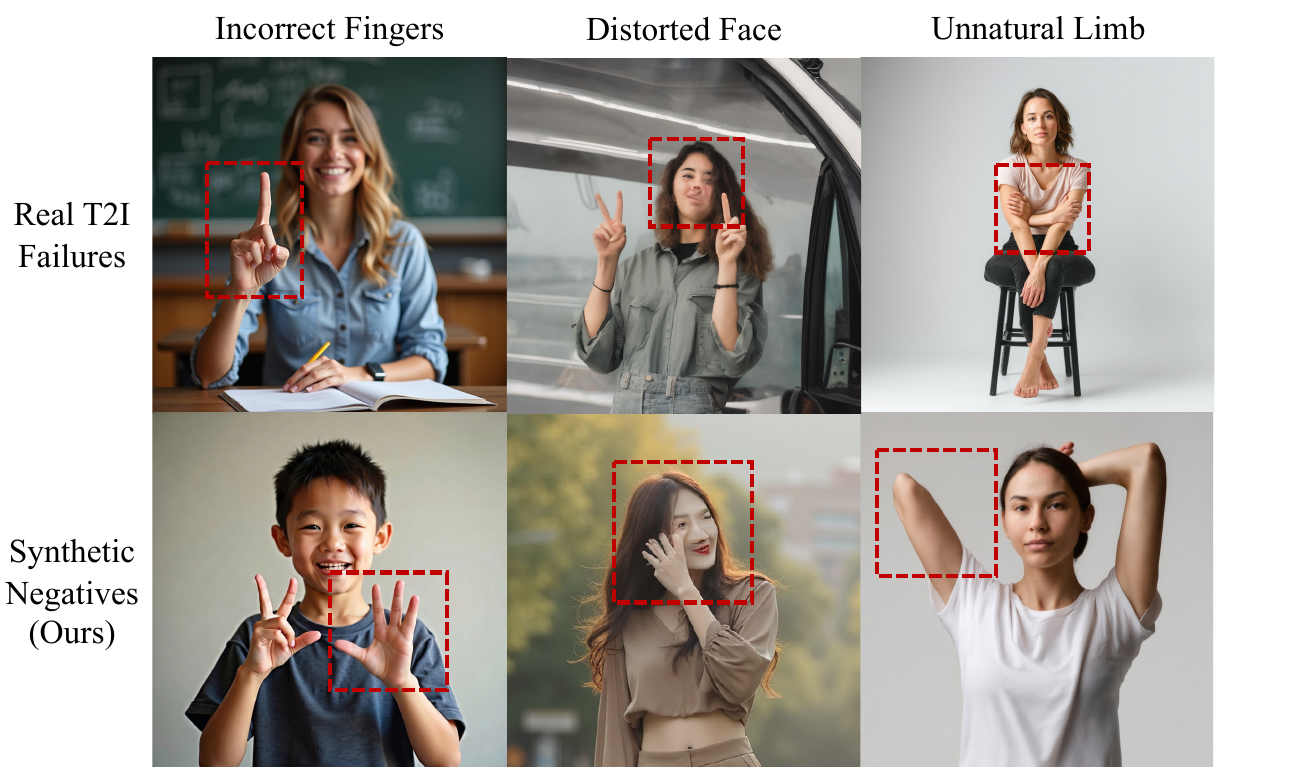}
  \caption{\textbf{Synthetic negatives capture representative anatomical failure patterns.} Each column pairs an anatomical failure observed in real T2I generation (top) with a synthetic negative from HAP (bottom) exhibiting the same error type. Highlighted regions indicate the anatomical artifacts.}
  \label{fig:real_modes}
  \vspace{-0mm}
\end{figure}

We further examine whether the constructed negatives resemble anatomical failures that arise naturally in text-to-image generation. As shown in Fig.~\ref{fig:real_modes}, untuned model outputs and synthetic negatives exhibit analogous defects, including extra or missing fingers and malformed facial structures. Beyond this visual correspondence, ASAP-DPO, trained on HAP pairs using the vanilla DPO objective, substantially reduces AER on unseen HAF-Bench prompts, as reported in Tab.~\ref{tab:main_results}. This downstream generalization indicates that HAP captures transferable anatomical failure patterns rather than arbitrary image corruptions. Together, the qualitative examples and quantitative improvements support the validity of HAP as anatomy-specific preference supervision, although the constructed negatives are not intended to cover the complete distribution of anatomical failures.

\section{HAF-Bench Evaluation Protocols}
\label{app:evaluation}

\subsection{Automated Evaluation}
Gemini-3-Pro serves as the automated judge under fixed anatomy-focused instructions shared across methods. In the absolute protocol, each generated image receives a Pass or Fail label as in Tab.~\ref{gemini_prompt}. AER is the fraction labeled Fail, with each image contributing one binary outcome irrespective of the number or severity of its artifacts. In the pairwise protocol, two images generated from the same prompt receive a Win, Tie, or Lose judgment from the evaluated model's perspective. RSI is computed as $(N_{\mathrm{win}}+N_{\mathrm{tie}})/(N_{\mathrm{lose}}+N_{\mathrm{tie}})$. The base model is the reference for each adapted model.

\subsection{Human Evaluation and Agreement}
Eleven human raters assess anatomical plausibility under protocols paralleling the automated evaluation. For absolute evaluation, they provide Yes/No judgments on anatomical artifacts. For pairwise evaluation, they select the more anatomically plausible image or a tie, with image order randomized. The main paper reports human-assessed AER and preference distributions for ASAP versus the corresponding base model on both backbones. To measure human--VLM agreement, we compare the human evaluation results from \cref{sec:human_evaluation} with the corresponding Gemini-3-Pro automated judgments across the entire HAF-Bench. We pool absolute Yes/No and pairwise Win/Lose/Tie decisions, counting agreement only when the labels match exactly. This comparison yields an overall agreement rate of 82\%. Although human and automated AER differ in value, both consistently favor ASAP. Human-assessed AER decreases from 45.2\% to 26.8\% for FLUX and from 97.5\% to 85.7\% for SDXL; automated AER decreases from 41.7\% to 28.9\% and from 87.3\% to 76.4\%, respectively. These results support a consistent improvement in anatomical plausibility.

\begin{table*}[t]
\centering
\caption{\textbf{Prompt for anatomical evaluation.}
The instruction supplied to Gemini-3-Pro for absolute evaluation. A ``Yes'' response is mapped to \textit{Fail}, whereas a ``No'' response is mapped to \textit{Pass}.}
\label{gemini_prompt}
\vspace{-0.4em}

\begin{tcolorbox}[
colback=gray!5!white,
colframe=gray!60!black,
boxrule=0.5pt,
arc=1pt,
left=3pt,
right=3pt,
top=3pt,
bottom=3pt,
boxsep=1pt,
fontupper=\scriptsize
]

\noindent\textbf{ROLE}\\
You are a Senior Visual QA Specialist specializing in Human Anatomy. Your specific objective is to identify \textbf{Severe and Obvious} anatomical artifacts in AI-generated human images.

\vspace{0.3em}
\noindent\textbf{TASK}\\
Conduct a general anatomical plausibility check on the person(s) in the image. Identify \textbf{glaring} anatomical errors in any visible part of the human body.

\vspace{0.3em}
\noindent\textbf{CRITICAL TOLERANCE SETTINGS}
\begin{enumerate}[leftmargin=*, noitemsep, topsep=0pt]
\item \textbf{IGNORE} minor artifacts, artistic stylization, and slight blur.
\item \textbf{IGNORE} background characters or heavily occluded body parts.
\item \textbf{ONLY FLAG} distortions that are physically impossible and immediately visible. If unsure, return ``No''.
\end{enumerate}

\vspace{0.3em}
\noindent\textbf{FAIL CRITERIA}\\
Analyze the image for the following defects. If any are found, return ``Yes'' (\textit{has distortion}).

\vspace{0.3em}
\noindent\textbf{1. Hands and Fingers}
\begin{itemize}[leftmargin=*, noitemsep, topsep=0pt]
\item \textbf{Obvious Polydactyly:} Explicitly visible extra fingers (\textit{e.g.}, six or more fingers on one hand).
\item \textbf{Unexplained Oligodactyly:} Fingers that are missing or cut off without a plausible cause. Do not flag fingers hidden by natural occlusion, such as when holding an object, forming a fist, appearing behind another body part, or being obscured by perspective. Flag only when the hand structure itself is clearly incomplete.
\item \textbf{Severe Malformation:} Fingers that branch incorrectly or exhibit claw-like or melted distortions.
\item \textbf{Fusion:} Fingers fused without visible separation or merged unnaturally into objects.
\end{itemize}

\vspace{0.3em}
\noindent\textbf{2. Face and Head}
\begin{itemize}[leftmargin=*, noitemsep, topsep=0pt]
\item \textbf{Severe Facial Collapse:} Facial features are merged, melted, or missing.
\item \textbf{Gross Misplacement:} Eyes, ears, or other facial features appear in anatomically implausible positions. Do not flag asymmetric pupils or an unusual gaze unless the eye structure itself is visibly malformed.
\end{itemize}

\vspace{0.3em}
\noindent\textbf{3. Body and Limbs}
\begin{itemize}[leftmargin=*, noitemsep, topsep=0pt]
\item \textbf{Physiologically Impossible Joints:} Limbs bend in impossible directions or attach to the body at structurally implausible angles.
\item \textbf{Surplus or Missing Limbs:} Extra limbs emerge from the body, required limbs are absent, or limbs appear detached.
\item \textbf{Severe Blending:} Limbs merge indistinguishably into clothing, furniture, or other objects.
\end{itemize}

\vspace{0.3em}
\noindent\textbf{OUTPUT RULES}
\begin{enumerate}[leftmargin=*, noitemsep, topsep=0pt]
\item Output only a valid JSON string without introductory text or Markdown formatting.
\item The JSON must contain exactly two keys: reasoning, containing a brief observation of the face, hands, and limbs, and has\_distortion,'' whose value must be either ``Yes'' or ``No''.
\end{enumerate}

\vspace{0.3em}
\noindent\textbf{OUTPUT EXAMPLES}\
Distortion found:
{
reasoning: [Briefly describe the severe anatomical errors observed],
has\_distortion: ``Yes''
}

No distortion found:
{
reasoning: [Briefly describe why the anatomical structures are plausible],
has\_distortion: ``No''
}

\end{tcolorbox}
\end{table*}

\section{Implementation Details}
\label{app:implementation}

\subsection{Training Settings}
\label{app:settings}

As introduced in the main paper, we use LoRA \citep{hu2022lora} with a rank of 32. All experiments are conducted on 8 NVIDIA A100 (80GB) GPUs. All images are processed at a resolution of $1024 \times 1024$.
For the optimization process, we employ the AdamW \citep{loshchilov2017decoupled} optimizer with a weight decay of 0.01. Both models are trained for 3,000 steps using an effective batch size of 16, accompanied by a linear learning rate warm-up of 200 steps.
Specifically for the FLUX \citep{flux2024} backbone, we apply a learning rate of $5e-6$, and set the spatial weighting scalar $\alpha = 10$ along with a regularization margin of $\tau = 0.01$ for our alignment objective. 
For SDXL \citep{rombach2022high} backbone, we apply a learning rate of $5e-5$, and set the spatial weighting scalar $\alpha = 10$ along with a regularization margin of $\tau = 0.002$ for our alignment objective.

\subsection{Baseline Adaptations}
\label{app:baselines}
\textit{SFT} fine-tunes each backbone using only the preferred images from HAP, without preference optimization. \textit{ASAP-DPO} instead applies vanilla DPO to exactly the same HAP pairs used by ASAP, but removes the localized preference gap and margin-bounded regression. Their comparison therefore isolates the contribution of preference supervision, while the comparison between ASAP-DPO and ASAP isolates the proposed alignment objective. The objective ablation in Tab.~\ref{tab:objective_ablation} further evaluates localized weighting and margin-bounded regression separately using the same training settings. The superscript $\dagger$ denotes baselines reproduced in our training framework rather than results or checkpoints taken directly from the original papers. Specifically, FairHuman$^\dagger$~\citep{wang2025fairhuman} and HG-DPO$^\dagger$~\citep{na2025boost} are trained using our data and the same backbone-specific optimization budget as the other methods. 
For FairHuman, the training code was publicly released, but the training data were unavailable. We retain its multi-objective supervision over global images, hands, and faces and apply it to our training samples. 
For HG-DPO, the original training data and official implementation of its data construction pipeline were unavailable to us. We therefore implement its easy-stage preference construction strategy using our data pool.

\subsection{Preference Construction Strategies}
\label{strategies}
To isolate the effect of preference-pair construction, all strategies start from the same prompt set and retain the same number of training pairs. The resulting datasets are used to train the corresponding backbone with vanilla DPO under identical optimization settings. The four strategies differ only in how their preferred and dispreferred samples are obtained: \textit{Standard Pairwise Ranking} independently generates two images for each prompt and assigns their preference order according to reward scoring. \textit{HG-DPO-Easy} selects highest and lowest reward samples from a larger candidate pool. \textit{Real-Anchor Localized Preferences} use real human images as preferred anchors and construct the corresponding dispreferred samples through localized anatomical degradation. \textit{Synthetic Localized Preferences} use human-verified synthetic images as preferred anchors and apply the same localized degradation principle to construct anatomically implausible negatives.

\section{Additional Benchmark Results}
To evaluate whether ASAP generalizes beyond our proposed HAF-Bench, we conduct an additional experiment on the publicly released Human-Art~\citep{ju2023human} validation split. We use all 1,250 text descriptions from its five natural-scene categories: Acrobatics, Cosplay, Dance, Drama, and Movie. These prompts are independently collected and were not involved in the construction of HAP or HAF-Bench. We compare FLUX with its ASAP-tuned model using identical prompts, seeds, and sampling settings. We report AER to evaluate anatomical plausibility, and HPSv3 to mesure overall image quality and prompt alignment.

\begin{table}[t]
\centering
\caption{\textbf{Evaluation on the Human-Art natural-scene validation subset.}
We use 1,250 text descriptions from five real-human categories.
AER evaluate anatomical plausibility, while HPSv3 measures
overall image quality and prompt alignment.}
\label{tab:humanart}
\small
\setlength{\tabcolsep}{7pt}{
\begin{tabular}{lcc}
\toprule
Method & AER $\downarrow$ & HPSv3 $\uparrow$ \\
\midrule
FLUX       & 0.223  & 10.140 \\
FLUX-ASAP (Ours) & 0.176  & 10.215 \\
\bottomrule
\end{tabular}}
\end{table}

\section{DPO Hyperparameter and Early-Stopping Analysis}
\label{early_stop}
\begin{figure}[t]
  \centering
  \makebox[\textwidth][c]
  {\includegraphics[width=1.0\linewidth]{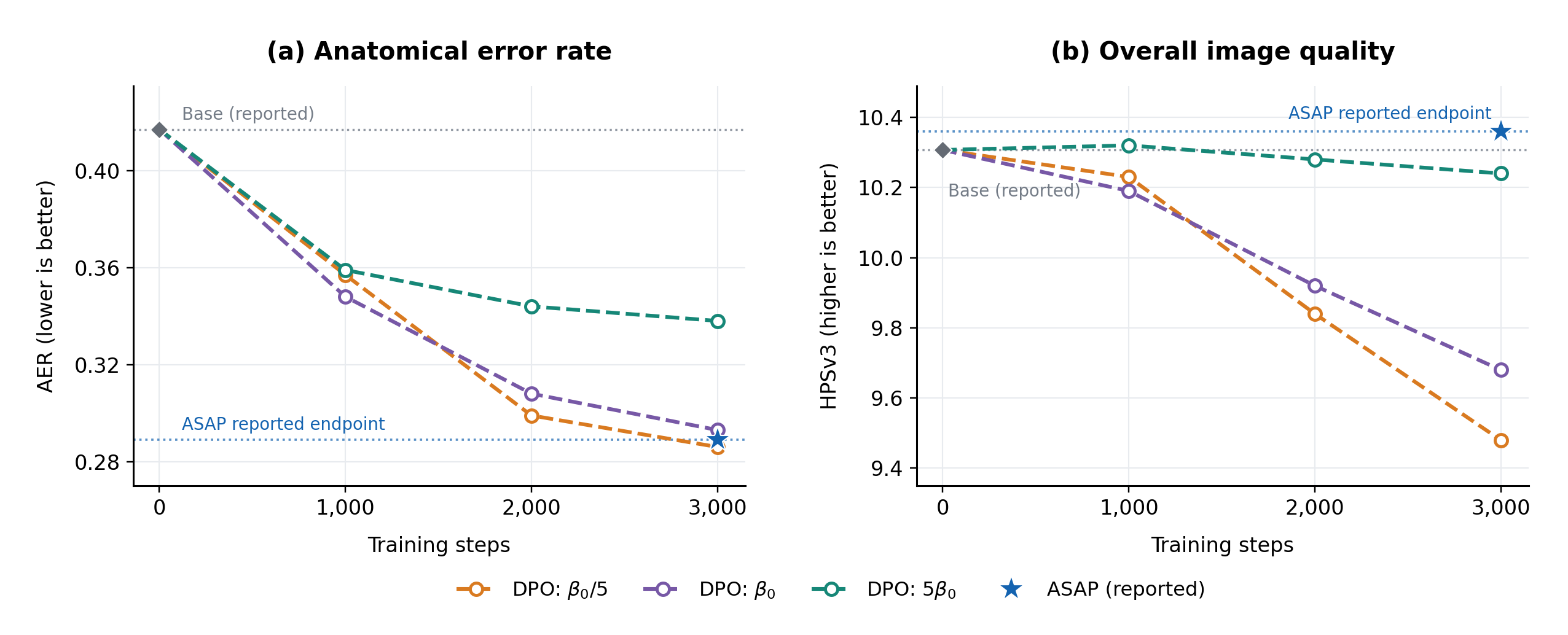}}
  \caption{\textbf{DPO tuning and early stopping for image-quality preservation.} We vary $\beta$ and training duration on FLUX.1-dev using the same HAP pairs. Adjusting $\beta$ or stopping early mitigates image-quality degradation but reduces anatomical gains. Within the evaluated configurations, ASAP achieves a better balance between anatomical plausibility and overall image quality.}
  \label{fig:early_stop}
  \vspace{0mm}
\end{figure}

To examine whether DPO's quality degradation can be mitigated through hyperparameter tuning or early stopping, we train FLUX on the same HAP pairs with $\beta \in \{\beta_0/5,\beta_0,5\beta_0\}$, where $\beta_0=100$ is our default DPO coefficient for FLUX. We evaluate checkpoints at 1,000, 2,000, and 3,000 steps under identical sampling conditions. As shown in \cref{fig:early_stop}, larger $\beta$ or earlier stopping generally preserves higher HPSv3 scores but yields less anatomical improvement. Further training reduces AER with diminishing gains, while overall image quality can deteriorate. Within the tested configurations, ASAP achieves a more favorable balance between anatomical plausibility and overall image quality.

\section{Limitations}
\label{app:limitations}

\begin{figure}[t]
  \centering
  \makebox[\textwidth][c]
  {\includegraphics[width=0.9\linewidth]{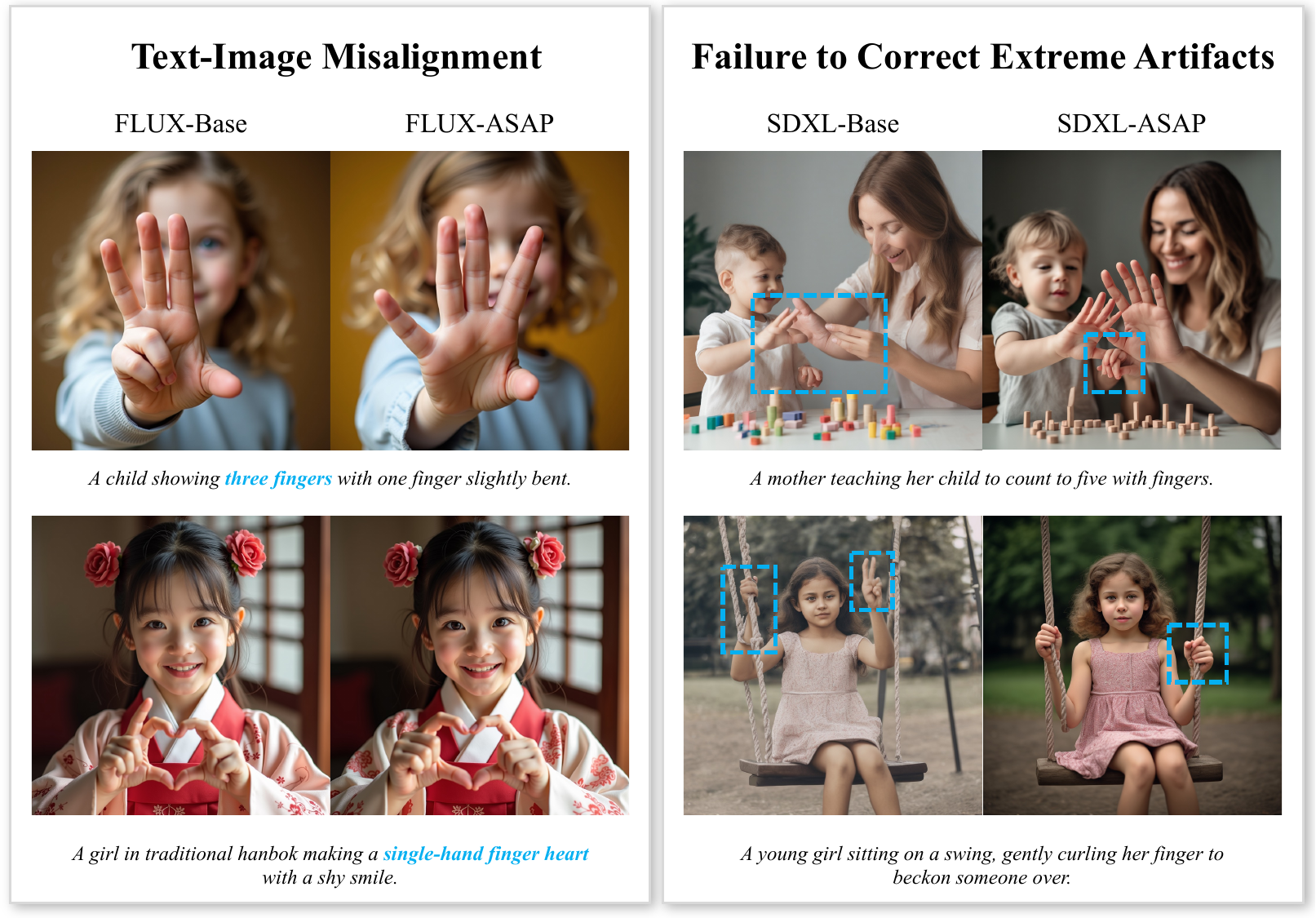}}
  \caption{  \textbf{Failure Case Analysis.} \textbf{(Left) Text-Image Misalignment:} While the ASAP-tuned models successfully generate anatomically plausible human images, they may occasionally fail to follow precise semantic constraints (\textit{e.g.}, numerical counting or specific gestures). \textbf{(Right) Failure to Extreme Artifacts:} In instances where the base model produces severe structural degradation when failing to resolve specific complex details, ASAP still enforces noticeable rectifications (\textit{e.g.}, separating severely fused fingers) but may fall short of achieving perfect anatomical plausibility.}
  \label{fig:failure}
  \vspace{-3mm}
\end{figure}

While our proposed ASAP framework significantly reduces anatomical errors, we observe two primary failure modes, both of which stem from the inherent limitations of the text-to-image foundation models, illustrated in Fig.~\ref{fig:failure}.

\paragraph{Text-Image Misalignment.} As illustrated in the left column, when prompts require precise numerical counting (\textit{e.g.}, "counting to five") or specific constraint-based gestures (\textit{e.g.}, "single-hand"), the ASAP-tuned models successfully generate highly realistic and anatomically plausible human images. However, they may occasionally fail to follow the precise semantic constraints of the text. This indicates that while ASAP effectively aligns the generative model toward anatomical correctness, it does not overcome the base model's inherent limitations in text-image compositional alignment.

\paragraph{Failure to Extreme Artifacts.}
We observe that, if the image generated by the base model with some seed contains extreme anatomical artifacts, the finetuned model may still generate artifacts in the corresponding human body parts. For instance, as demonstrated in the right column, when the base model produces severe anatomical artifacts across extensive regions, ASAP still enforces noticeable rectifications (\textit{e.g.}, separating fused fingers), but frequently falls short of achieving perfect anatomical plausibility. This indicates that, as a fine-grained alignment strategy, ASAP fundamentally requires a minimally plausible anatomical foundation from the base generative model.

\end{document}